\documentclass[reprint, cha]{revtex4-1}

\usepackage{amssymb}
\usepackage[english]{babel}    % para texto em Inglês
\usepackage[dvips]{graphics}
\usepackage{subfigure}
\usepackage[pdftex]{graphicx}
\usepackage{multirow}
\usepackage{rotating}
\usepackage{picinpar}
\usepackage{color}
\usepackage{lineno}
\usepackage{listings}

\begin{document}

\title{Texture analysis using volume-radius fractal dimension}

\author{Andr\'e. R. Backes}
 	     \email{backes@facom.ufu.br}
\affiliation{Faculdade de Computa\c{c}\~{a}o, Universidade Federal de Uberl\^{a}ndia, Av. Jo\~{a}o Naves de \'Avila, 2121, 38408-100, Uberl\^{a}ndia, MG, Brasil} 

\author{Odemir M. Bruno}
              \email{bruno@ifsc.usp.br}
\affiliation{Instituto de F\'{i}sica de S\~{a}o Carlos (IFSC), Universidade de S\~{a}o Paulo, Av.   Trabalhador S\~{a}o Carlense, 400, 13560-970, S\~{a}o Carlos, SP, Brasil}

\date{\today}

\begin{abstract}
Texture plays an important role in computer vision. 
It is one of the most important visual attributes used in image analysis, once it provides information about pixel organization at different regions of the image. 
This paper presents a novel approach for texture characterization, based on complexity analysis. 
The proposed approach expands the idea of the Mass-radius fractal dimension, a method originally developed for shape analysis, to a set of coordinates in 3D-space that represents the texture under analysis in a signature able to characterize efficiently different texture classes in terms of complexity. 
An experiment using images from the Brodatz album illustrates the method performance.
\end{abstract}


\maketitle

%% main text
\section{Introduction}
Texture is a visual attribute that performs an important role in computer vision, image analysis and pattern recognition. There are a lot of applications using textures in different areas of knowledge, ranging from medical images \cite{wu_1992}, passing 
by remote sensing \cite{yang_1998}, analysis of geological images \cite{Heidelbach_2000}, 
etc.

The full definition of texture is a complex task. Indeed, there is no formal definition in the literature that is capable of explaining it completely. This occurs, due to the nature of the texture that can be modeled in different ways. Texture can be formed by simple repetitions of set of pixels or simple patterns, but it can also be formed by complex arrangements. These arrangements can be constituted by tiles of natural patterns such as leaves, rocks, clouds or even for more abstract patterns. In fact, even the absence of patters can characterize a texture (e.g., a region formed by noise in an image). If, on one hand, texture is very difficult to be formally defined, on the other hand, the importance of the attribute and its application has been motivated the development of many algorithms and methods. Over the years, many approaches have been proposed to describe texture patterns: second-order statistics \cite{haralick-1979,journals/pr/MurinoOP98}, spectral analysis \cite{journals/paa/ShenB06,journals/pr/BianconiF07,jain-1991,daugman95gabor,journals/pami/ManjunathM96,journals/pami/AzencottWY97,bb28041}, wavelet packets \cite{journals/eswa/SengurTI07,unser95} and fractal dimension \cite{BackesB08,journals/ijprai/ChenB99,books/tricot}.

This paper presents a novel approach for texture characterization, based on fractal analysis. The proposed approach expands the methodology of the Mass-radius fractal dimension. The Mass-radius fractal dimension was originally developed to deal with binary images and shape analysis. The proposed method considers the pixels of an image as a set of coordinates in 3D-space, where the z-axis is the pixel intensity. In this way, it estimates the fractal dimension of the surface of the image and, consequently, it is capable of dealing with textures. Besides the extension of the Mass-radius fractal dimension method, the paper approach uses a signature, obtained by a vector calculated by the fractal dimension, to characterize textures. The method is described in detail and an experiment using images from Brodatz album shows the method performance. The proposed method is compared with popular texture ones.

\section{Proposed Approach}
\label{sec:approach}

In this work, a novel approach for texture analysis is proposed. The approach is based on the mass-radius method for shape complexity analysis \cite{Fernandez2001309,Caserta1995133,journals/bioinformatics/LandiniR93}. This method consists of covering the shape with circles of radius $r$ and to compute the amount of shape that is intercepted by the circle as the radius increases. 

Consider an image texture as a set of coordinates $S$. Each texture pixel is represented as a triple $s = (y,x,z)$, $s \in S$, where $y$ and $x$ are the Cartesian coordinates of the pixel at the original image and $z$ is the gray-level associated to the pixel $(y,x)$. Note that now an image is represented as a set of points in a 3D-space. Thus, the circle employed in the original method is replaced by a sphere of radius $r$ and the amount of points $s \in S$ intercepted by the sphere is computed.

One important step in the method is to select the total number of spheres that will be employed to sample the texture complexity. Each sphere is centered at a specific point $s_i \in S$, $u = 1,2,\dots,N$, randomly chosen. So, the number of points intercepted by a sphere of radius $r$, $V_{i}(r)$, is defined as:

\begin{equation}V_{i}(r) = \left| \left\{ s_i \in S | \exists s \in S: \left| s - s_i \right| \leq r \right\} \right|,\end{equation}
where $s_i$ is a point in $S$ which dists $r$ or less from $s$. For $N$ spheres, we consider the occupied volume $V(r)$ as

\begin{equation}V(r) = \frac{1}{N} \sum_{i=1:N} V_{i}(r)\end{equation}

From occupied volume $V(r)$, the fractal dimension $D$ is estimated as
\begin{equation} D = \lim_{r \to 0} \frac{\log{V(r)}}{\log{r}}.\end{equation}

\section{Texture signature}
\label{sec:signature}
From log-log curve computed from proposed approach, the fractal dimension can be easily estimated by applying linear regression over the curve $\log{r} \times \log{V(r)}$, where the resulting line presents angular coefficient $\alpha$ and $D = \alpha$ is the estimated fractal dimension. 

However, a single non-integer value may not be suitable to represent all complexity and self-similarity present in the image. In fact, if we analyze the computed log-log curve we may note that it presents considerable information along the scales that are lost during the process of linear regression.

\begin{figure}[htbp]
	\centering
	\includegraphics[width=\columnwidth]{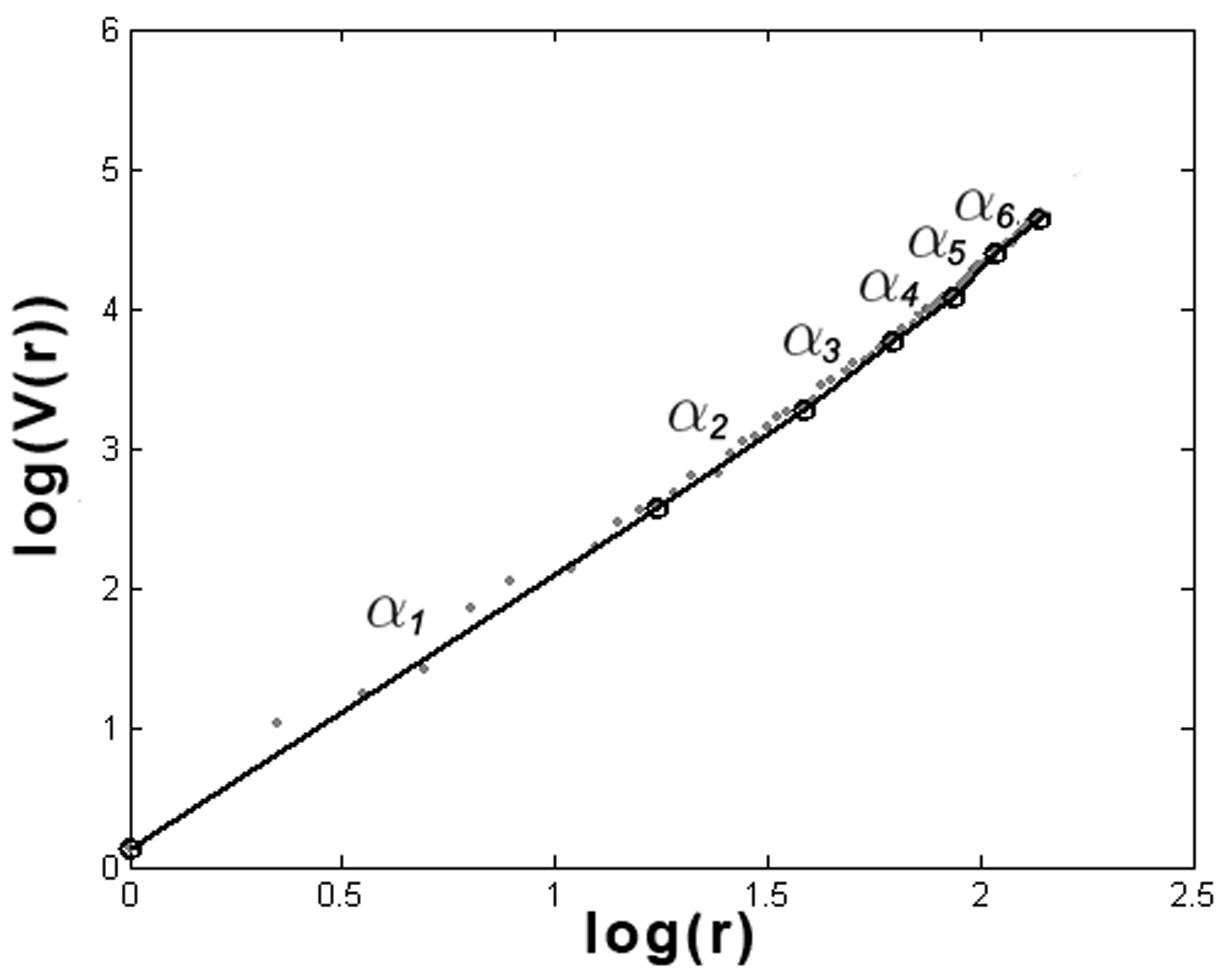}	
	\caption{Example of texture signature computed for $M = 10$, resulting in $k = 8$ line segments. Log-log curve computed for $r = 10$.}
	\label{fig:assinatura}
\end{figure}

Thus, we propose to compute the linear regression at different sections of the log-log curve. Each linear regression is computed for a section of the curve composed by $M$ points in sequence and one single point is not allowed to belong to two different curve sections (Figure \ref{fig:assinatura}). 

As a result, a vector $\vec{\varphi} = \left\{ \alpha_1, \alpha_2, \dots, \alpha_k, \right\}$ capable of describing the complexity changes at different portions of the log-log curve is yielded, where $k$ is the number of line segments computed, thus providing a more efficient texture characterization.

\section{Experiments}
\label{sec:experiments}
\begin{figure}[htbp]
	\centering
	\includegraphics[width=\columnwidth]{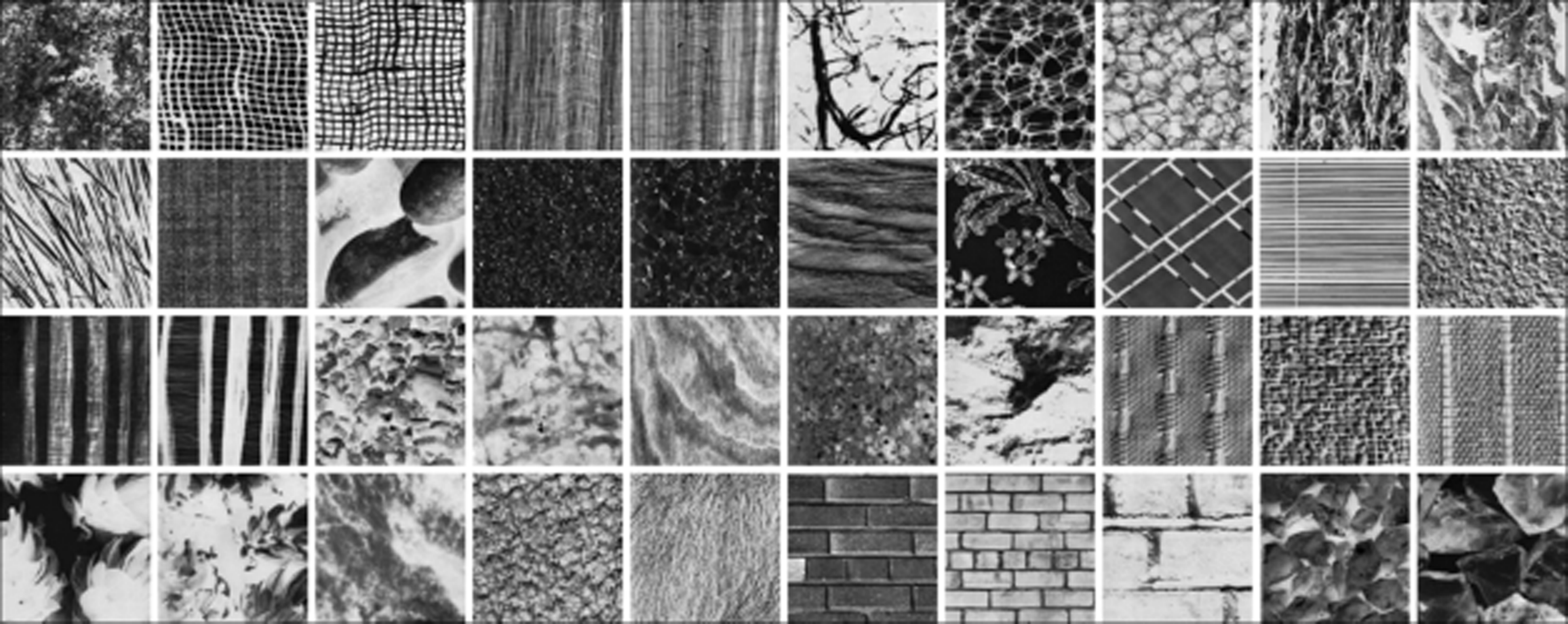}
	\caption{Example of each Brodatz texture class considered.}
	\label{fig:brodatz}
\end{figure}

The proposed approach was evaluated considering images collected from Brodatz album \cite{brodatz-1966}. These images were selected once they are widely employed by literature as benchmark for texture analysis methods in computer vision and image processing applications. Each image considered has 200 $\times$ 200 pixels of size, with 256 gray levels. The image set used contains 400 images grouped into 40 Brodatz classes, with 10 samples each. Figure \ref{fig:brodatz} presents one example of each texture class considered in the experiment.

The proposed signature was computed for each texture and analysis step was carried out applying a Linear Discriminant Analysis (LDA) \cite{everitt-2001,fukunaga-1990}. The LDA is a supervised method which enables us to find a feature space where the distribution of classes presents good discriminative properties. Descriptors are considered "good" when the variance between classes is larger than the variance within classes in this feature space. Leave-one-out cross-validation scheme was also employed during the analysis.

\section{Results and Discussion}
\label{sec:results}

\begin{figure}[!htbp]
	\centering
	\includegraphics[width=\columnwidth]{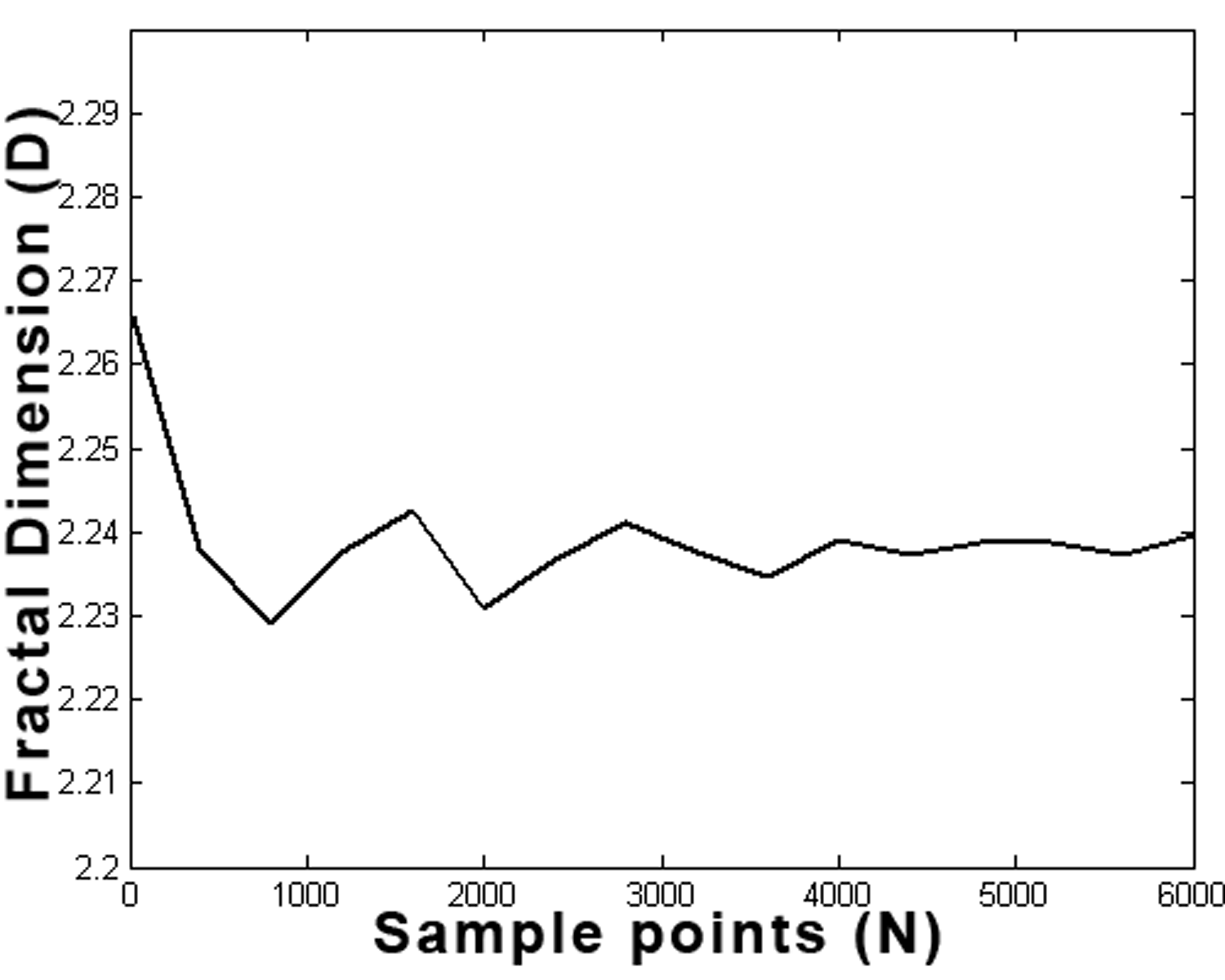}
	\caption{The fractal dimension $D$ as a function of the number of points $N$ considered for texture sampling.}
	\label{fig:amostragemDF}
\end{figure}

An important issue from the proposed approach that claims for attention refers to the number of spheres $N$ used by the method to sample the texture pattern under analysis. Each sphere is centered at random over the texture and a small number of spheres may produce an unsuitable texture sampling. This may result in an underestimated fractal dimension value or even in different fractal dimensions for different executions of the method over a same texture sample. Otherwise, after a given number of spheres, the texture is over sampled, i.e., no relevant information is add to the log-log curve by each additional sphere. Figure \ref{fig:amostragemDF} shows the fractal dimension value $D$ estimated for a given texture sample according to the number of spheres $N$ used during the sampling step. In this experiment, for each value of $N$, the fractal dimension $D$ was estimated 30 times and its average computed. As a result, we note that fractal dimension is stable for $N \geq 4000$, which corresponds to select 10\% of the texture pixels during its sampling step.

\begin{figure}[!htbp]
	\centering
	\includegraphics[width=\columnwidth]{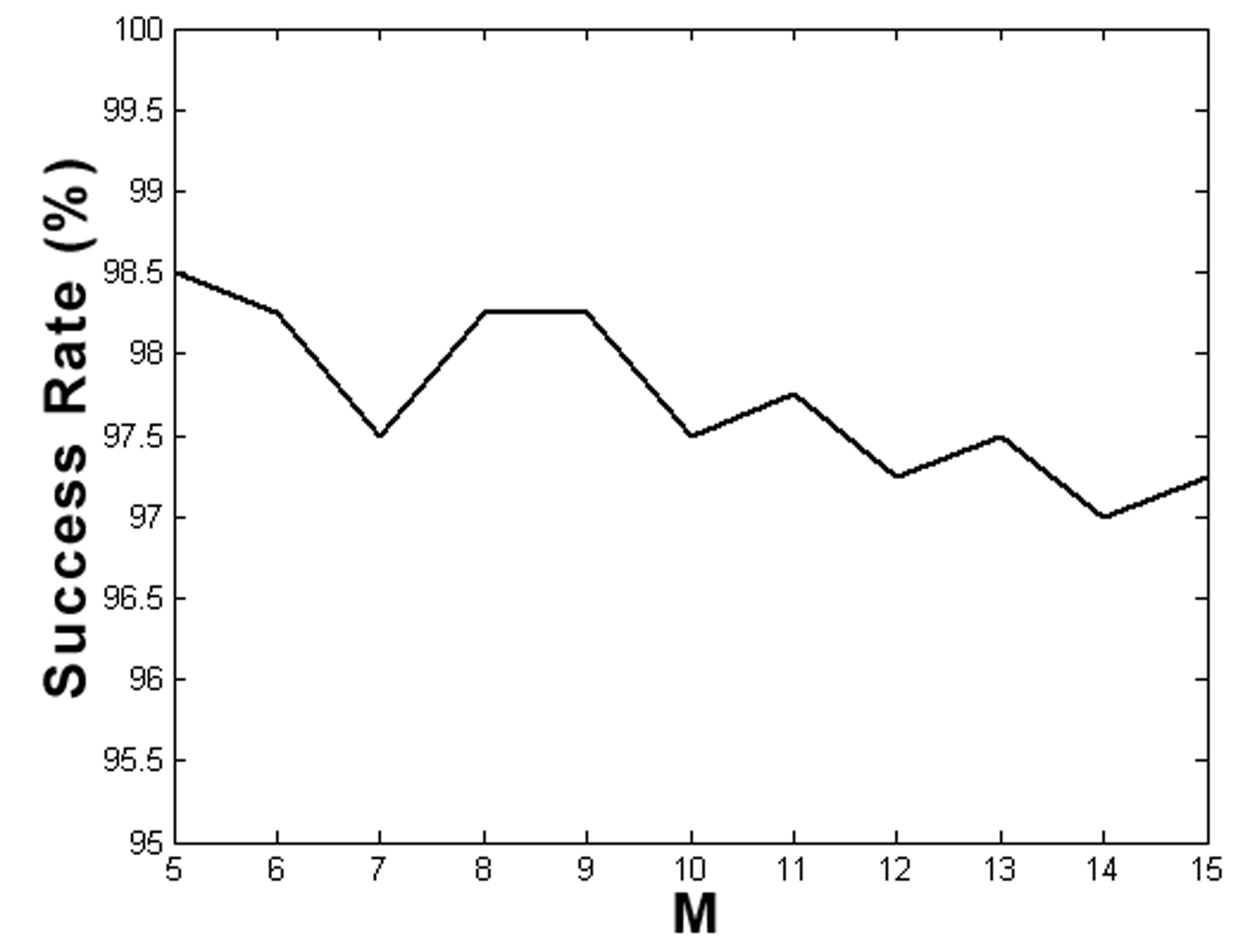}
	\caption{Success rate as a function of the slope interval ($M$) considered. Best classification (98.50 \%) is achieved when using $M = 10$.}
	\label{fig:espacamento}
\end{figure}

Another important parameter of the method is the number of points $M$ used to compute the texture signature. Figure \ref{fig:espacamento} shows the success rate of the proposed approach according to the number of point $M$ used to compute each line segment during the texture signature making process. In the experiments, a maximum radius $r = 20$ was considered, thus resulting in a log-log curve containing $335$ points. Each point in the curve corresponds to a radius value at interval $\left[0,radius\right]$ in the discreet 3D-space. According to the length of the line segments, different sections of the log-log curve are selected, so emphasizing details at different resolutions. As the line segment increases, more information is used to compute a single angular coefficient. We have that different oscillations in the log-log curve are now represented by the same angular coefficient. These oscillations are due to the volume of the spheres do not increase equally in all texture points sampled. Small variations in the texture pattern disturb how the influence volume $V(r)$ increases, and this makes $V(r)$ very sensitive to structural changes on texture patterns. Thus, an increase in the line segment size tends to decrease the relevance of the details present in that section of the curve. Moreover, longer line segments produce a smaller set of linear coefficients and, as a consequence, a less discriminative texture signature. In fact, a subtle decrease in success rate is perceived as $M$ increase and the best result (98.50 \%) is found when $M = 5$ is considered.

\begin{table*}[!htbp]
	%\scriptsize
	\centering
		\begin{tabular}{ccc}
			\hline
			Method & Images correctly classified & Success rate (\%)\\			 
			\hline									
			Co-occurrence matrices & 330	& 82.50\\
			Fourier descriptors & 351	& 87.75\\
			Gabor Filters & 381 & 95.25\\
			Proposed Method & 394 & 98.50\\
			\hline			
		\end{tabular}
	\caption{Comparison results for different texture methods.}
	\label{tab:Results_method}
\end{table*}

Results yielded by different texture analysis methods are presented in Table \ref{tab:Results_method}. The methods considered for comparison are: Fourier descriptors \cite{journals/pami/AzencottWY97}, Co-occurrence matrices \cite{haralick-1979} and Gabor filters \cite{jain-1991,daugman95gabor,Idrissa-2002}. A brief description of the methods is presented as follows:

\textit{Fourier descriptors}: it is a set containing the energy of the 99 most meaningful coefficients of the Fourier Transform applied over the image. Each coefficient represents the sum of the spectrum absolute values from a given radial distance from the center transformation.

\textit{Co-occurrence matrices}: they represent the joint probability distributions between the gray-levels of pairs of pixels at a given orientation and distance. Energy and entropy were computed from non-symmetric matrices obtained for distances of 1 and 2 pixels with angles of $-45\,^{\circ}$, $0\,^{\circ}$, $45\,^{\circ}$, $90\,^{\circ}$, totalizing 16 descriptors. 

\textit{Gabor filters}: an input image is convolved by a family of filter, where each filter is a bi-dimensional gaussian function moduled with an oriented sinusoid in a determined frequency and direction. In this paper, 16 filters (4 rotation filter and 4 scale filters), with lower and upper frequencies equal to 0.01 and 0.3, respectively, were employed. Energy from the resulting images was used as its descriptors.

The proposed approach performs texture analysis directly over texture pixels, i.e., no transformation is applied over the image pixels. However, its result overcomes the ones from traditional texture analysis methods, such as Fourier descriptors and Gabor filters. These methods employ more complexes and sophisticated computing than the proposed approach, and this contributes to validate our approach as a feasible texture descriptor.

\section{Conclusion}
\label{sec:conclusion}

This paper presented a novel approach for texture discrimination using complexity analysis. The proposed approach is based on the idea of the mass-radius method, which is used in literature to compute the fractal dimension of shapes. By considering texture as a set of coordinates in 3D-space, the original method is easily expanded from shape analysis to texture analysis, thus enabling us to estimate the fractal dimension of a texture pattern. An experiment using the texture signature computed using the proposed approach and linear discriminant analysis to classify texture samples extracted from Brodatz album was performed. Results show that the method presents great potential to be used in texture identification/classification tasks.

\section*{Acknowledgments}

A.R.B. acknowledges support from FAPESP (2006/54367-9).
O.M.B. acknowledges support from CNPq (306628/2007-4 and 484474/2007-3).

%\bibliographystyle{elsarticle-num}
%\bibliography{Bibliografia}

%\begin{thebibliography}{00}
%
%%% \bibitem{label}
%%% Text of bibliographic item
%
%\bibitem{}
%
%\end{thebibliography}

\end{document}